\documentclass[letterpaper]{article} 
\usepackage{aaai24}
\usepackage{times}  
\usepackage{helvet}  
\usepackage{courier}  
\usepackage[hyphens]{url}  
\usepackage{graphicx} 
\usepackage{natbib}  
\usepackage{caption} 
\usepackage{algorithm}
\usepackage{algorithmic}
\usepackage{amsmath}
\usepackage{amsthm,amssymb}
\usepackage{mathrsfs}
\usepackage{bm}

\usepackage{svg}

\def\mb{\mathbf}
\def\mbb{\mathbb}
\def\mc{\mathcal}

\def\ie{\textit{i.e.}}

\newtheorem{definition}{Definition}

\usepackage{newfloat}
\usepackage{listings}
\DeclareCaptionStyle{ruled}{labelfont=normalfont,labelsep=colon,strut=off} 
\floatstyle{ruled}
\newfloat{listing}{tb}{lst}{}
\floatname{listing}{Listing}
\nocopyright 

\title{Long-Tailed Learning as Multi-Objective Optimization}
\author{
    Weiqi Li\textsuperscript{\rm 1},
    Fan Lyu\textsuperscript{\rm 1}\textsuperscript{\rm 2}
    Fanhua Shang\textsuperscript{\rm 1},
    Liang Wan\textsuperscript{\rm 1},
    Wei Feng\textsuperscript{\rm 1}\thanks{Corresponding author}
}
\affiliations{
    \textsuperscript{\rm 1}College of Intelligence and Computing, Tianjin University\\
	\textsuperscript{\rm 2}CRIPAC, MAIS, CASIA\\
	\{wicky, fhshang, lwan\}@tju.edu.cn, fan.lyu@cripac.ia.ac.cn, wfeng@ieee.org

}

\begin{document}

\maketitle

\begin{abstract}
Real-world data is extremely imbalanced and presents a long-tailed distribution, resulting in models that are biased towards classes with sufficient samples and perform poorly on rare classes. Recent methods propose to rebalance classes but they undertake the seesaw dilemma (what is increasing performance on tail classes
may decrease that of head classes, and vice versa). 
In this paper, we argue that the seesaw dilemma is derived from gradient imbalance of different classes, in which gradients of inappropriate classes are set to important for updating, thus are prone to overcompensation or undercompensation on tail classes.
To achieve ideal compensation, we formulate the long-tailed recognition as an multi-objective optimization problem, which fairly respects the contributions of head and tail classes simultaneously. For efficiency, we propose a Gradient-Balancing Grouping (GBG) strategy to gather the classes with similar gradient directions, thus approximately make every update under a Pareto descent direction. Our GBG method drives classes with similar gradient directions to form more representative gradient and provide ideal compensation to the tail classes. Moreover, We conduct extensive experiments on commonly used benchmarks in long-tailed learning and demonstrate the superiority of our method over existing SOTA methods.
\end{abstract}

\section{Introduction}

Deep learning has made significant progress and been widely applied in many applications~\cite{li2022nested,tan2020equalization}. Most of these excellent achievements rely on large and relatively balanced datasets, such as ImageNet~\cite{deng2009imagenet} and MS-COCO~\cite{lin2014microsoft}. However, real-world data is often extremely imbalanced, presenting a long-tailed distribution.
Training on long-tailed data usually results in serious bias towards classes with sufficient samples ({head classes}) and performs poorly on rare classes ({tail classes}), giving rise to the field of long-tailed learning.

\begin{figure}[t!]
	\centering
	\includegraphics[width=1.0\linewidth]{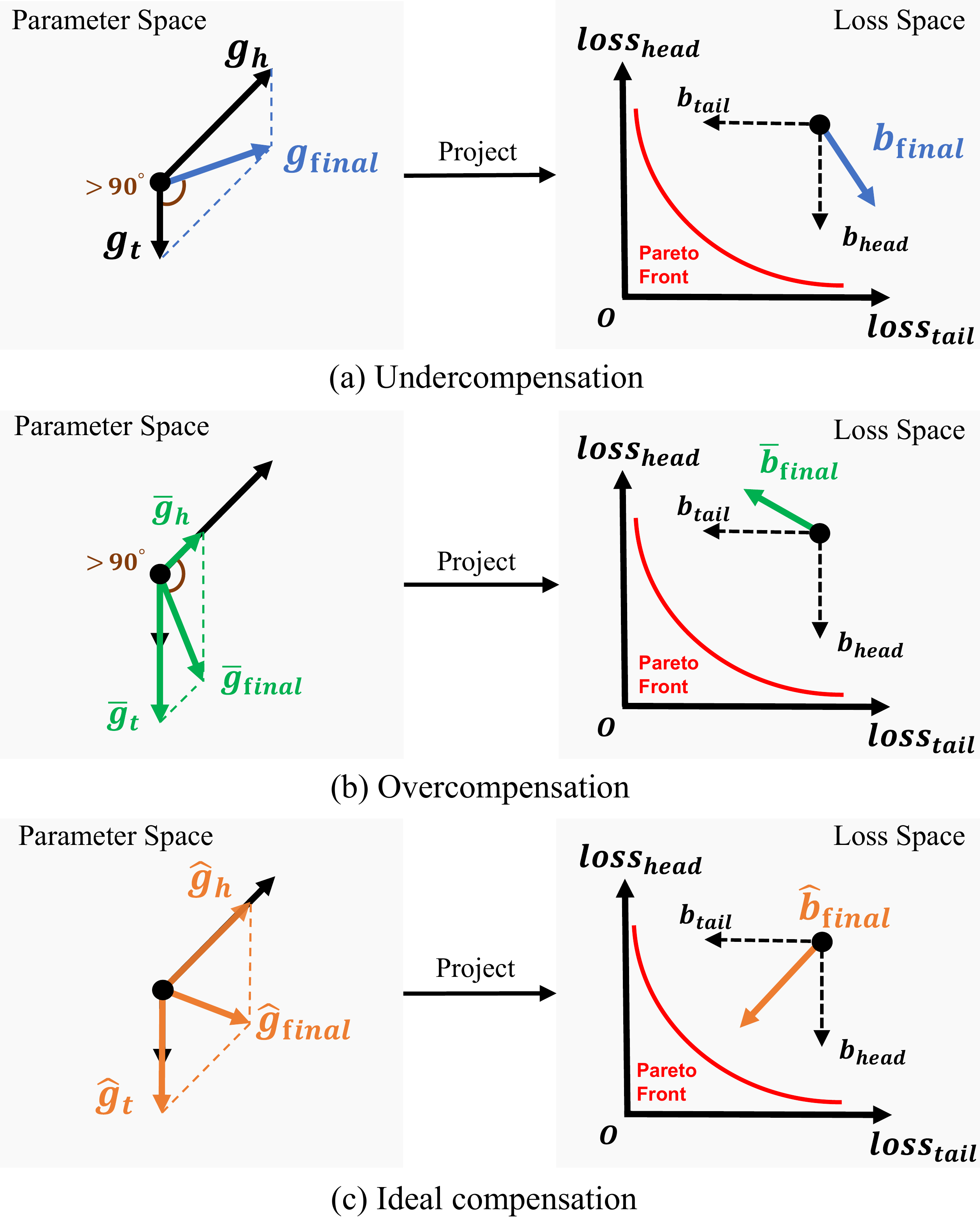}
	\vspace{-0.2in}
	\caption{Three gradient compensation scenarios in reweighting for two-class imbalance training: (a) undercompensation; (b) overcompensation; (c) ideal compensation.
        By optimizing two classes simultaneously at each step, an ideal gradient should step towards the Pareto front without harming both classes, which means the loss descent ($b_\text{final}$) should be suited between two class-independent loss descent directions ($b_{head}$ and $b_{tail}$).
 }

	\label{fig:seesaw}
	\vspace{-0.2in}
\end{figure}

To address the problem of learning in long-tailed distribution, recent progress on long-tailed learning can be categorized into three groups.
First, the class-rebalancing methods~\cite{zhang2021deep} increase the importance of tail classes via resampling or reweighting directly.
Second, the decoupling methods~\cite{zhou2020bbn} use a two-stage training scheme to balance the classifier after observing from a pre-training phase.
Third, the representing methods~\cite{cao2019learning} design specific loss functions to achieve inter-class sparsity and a more balanced feature distribution.
To sum up, the key consensus of these methods is to improve the importance of tail classes in long-tailed training. 
However, the existing rebalancing~\cite{zhang2021deep,kang2019decoupling,shu2019meta} methods aiming to increase the importance of tail-class gradients, may suffer from the \emph{seesaw dilemma}.
{That is, to increase performance on tail classes may decrease that of head classes, and vice versa.}




In this paper, we study the seesaw dilemma in the perspective of gradient imbalance in long-tailed learning, and we observe that the tail-class gradients are suppressed by those of head classes.
Under this observation, inappropriate weighting scheme may lead to the \emph{overcompensation} or \emph{undercompensation} on gradient of tail classes.
In general, undercompensation refers to a bias towards head class learning and overcompensation refers to the over-bias towards learning tail classes.
Taking an imbalanced two-class classification as an example, we illustrate the effects of different compensations in Fig.~\ref{fig:seesaw}. 
By projecting from parameter space to loss space, we find that undercompensation may result in insufficient learning (Fig.~\ref{fig:seesaw}(a)) for tail classes, while overcompensation may hinder the learning of head classes (Fig.~\ref{fig:seesaw}(b)).
Ideally, \textit{a feasible compensation to the gradients in a long-tailed problem should maintain a Pareto descent direction~\cite{harada2006local}, which should never damage any classes in the imbalanced distribution}, demonstrated in Fig.~\ref{fig:seesaw}(c).


To achieve feasible compensations in the seesaw dilemma, we propose to formulate the long-tailed learning into a multi-objective optimization problem (MOO), where each class holds its class-level training empirical loss. 
In this way, our goal is to find a compromising gradient that not damage any of these losses at each update from a set of class-level gradients.
Furthermore, it is impractical to extract gradients {for every class independently} in each mini-batch training because of two reasons.
On one hand, more classes lead to 
{more computation time and may also cause the out-of-memory problem.}
On the other hand, a limited batch size can not guarantee {the access to each class} especially tail classes leading to incompleteness of objectives in each optimization step.

Accordingly, in this paper, we develop a Gradient-Balancing Grouping (GBG) algorithm to present a batch-level gradient balance in long-tailed learning.
Specifically, we first compute the gradients of all classes and obtain the gradient similarity between classes to build an similarity matrix. 
Then, we learn to group the classes with similar gradients according to the similarity matrix.
To obtained a balanced gradient to guarantee the Pareto descent, inspired by the classic multi-gradient descent algorithm~\cite{sener2018multi}, we take the bundled gradients from each groups as a min-norm optimization, which can be solved easily via quadratic programming. 
\par Our main contributions are three-fold:
\begin{itemize}
\vspace{-5px}
\setlength{\itemsep}{0pt}
\setlength{\parsep}{0pt}
\setlength{\parskip}{0pt}
\item To the best of our knowledge, it is the first time that we formulate long-tailed recognition as a multi-objective optimization problem, to {address the seesaw dilemma for the head and tail classes} in previous methods.
\item We propose a grouping method based on gradient similarity to solve the multi-objective optimization efficiently without compromising accuracy.
\item Our method has been validated to outperform state-of-the-art works on the broadly used benchmarks including CIFAR10/100-LT, ImageNet-LT and INaturalist2018, {which demonstrate its capability in solving long-tailed problems efficiently.}
\end{itemize}

\section{Related Work}


\noindent\textbf{Long-tailed Learning via Class-Rebalancing.}  Class-Rebalancing includes resampling and reweighting. Resampling strategies aim to attain a balanced training data distribution. They use 
over-sampling~\cite{buda2018systematic} to enlarge instance number of tail classes 
or use under-sampling~\cite{he2009learning} to decrease that of head classes. But they afford risks of overfitting tail classes or impairing model generalization.
Class-Reweighting methods assign weights to the loss functions of each class that are negatively correlated with their sample sizes, aiming to balance the gradient contribution of different classes~\cite{jamal2020rethinking}.
However, inappropriate weights used in reweighting methods may cause problems such as underfitting or overfitting of the model. 

\noindent\textbf{Long-tailed Learning via Grouping Strategy.} Grouping strategies decompose long-tailed classification problem into a multi-task problem or a multi-level recognition problem by grouping the label sets according to certain rules~\cite{yang2022survey} such as grouping based on instance numbers of classes~\cite{li2020overcoming}.
Though current grouping strategies can avoid tail categories being suppressed to some extent, they could not solve the problem of knowledge interaction blocking between different groups~\cite{yang2022survey}. 

\noindent\textbf{Multi-Objective Optimization in Deep Learning.} 
Multi-Objective Optimization (MOO) refers to optimizing multiple objective functions which may be conflicting in optimization problems. The target of MOO is to find a set of optimal solutions that can simultaneously optimize multiple objectives~\cite{lyu2023measuring}. MOO can be applied to fields that require simultaneously optimizing multiple targets such as multi-task learning~\cite{sener2018multi,lyu2021multi,chen2023multi} and recommendation systems~\cite{geng2015nnia}. In this paper, we use MOO to balance the learning of head classes and tail classes.

\section{The Proposed Method}

\subsection{Gradient Imbalance Problem in LT Learning}

\label{sec:grad_imblance}


{Let $\mc{D}=\{(x_i,y_i),\cdots,(x_N,y_N)\}$ denotes a long-tailed training set,}
with totally ${N}$ samples and ${K}$ classes.
Long-tailed classification aims to learn a function $f\left(\boldsymbol{\theta}\right)$ with parameters $\mathbf{\theta}$ to predict each test sample correctly.
For a data point $(x_i,y_i)$, $x _i$ represents the \textit{i}-th data point in the training set and $y_i$ represent its ground-truth label. 
Usually, the model will be trained using an empirical risk loss as follows:
\begin{equation}
    {L}\left(\mathbf{x},\mathbf{y}\right) = \frac{1}{{N}}\sum_{i=1}^{{N}}{L}\left(x_i,y_i\right) =  -\frac{1}{N}\sum_{i=1}^{{N}}\log\left(\frac{e^{z_{y_i}}}{\sum_{j=1}^{K}e^{z_j}}\right),
    \label{eq:CE_batch}
\end{equation}
where $z_j$ is the predicted logit of class $j$ and $z_{y_i}$  is the logit of the corresponding ground-truth class.

To explore the gradient imbalance in long-tailed learning, we split the training loss into head and tail losses as follows:
\begin{equation}
    {L}\left(\mathbf{x},\mathbf{y}\right) 
    =\frac{1}{{N}}\left[\sum\limits_{i=1}^{{N}_\text{tail}}{{L}\left(x_i,y_i\right)} + \sum\limits_{j=1}^{{N}_\text{head}}{L}\left(x_i,y_i\right)\right],
    \label{eq:loss_ht_split}
\end{equation}
where we use ${N}_\text{head}$ and ${N}_\text{tail}$ to represent the numbers of head-class samples and tail-class samples in a mini-batch.
Thus, the gradient of parameter $\mathbf{\theta}$ can be denoted as:
\begin{equation}
   \setlength{\abovedisplayskip}{5pt}
\setlength{\belowdisplayskip}{5pt}
\begin{aligned}
        \nabla_{\boldsymbol{\theta}} =\frac{\partial{{L}\left(\boldsymbol{x},\boldsymbol{y}\right)}}{\partial{\boldsymbol{\theta}}} =\nabla_{\boldsymbol{\theta}}^\text{tail}+\nabla_{\boldsymbol{\theta}}^\text{head},
        \end{aligned}
\end{equation}
where $\nabla_{\boldsymbol{\theta}}^\text{tail}$ and $\nabla_{\boldsymbol{\theta}}^\text{head}$ are the gradients of $\boldsymbol{\theta}$ generated by the tail-class and head-class instances in the mini-batch. 

\begin{figure}[t!]
	\centering
	\includegraphics[width=1.0\linewidth]{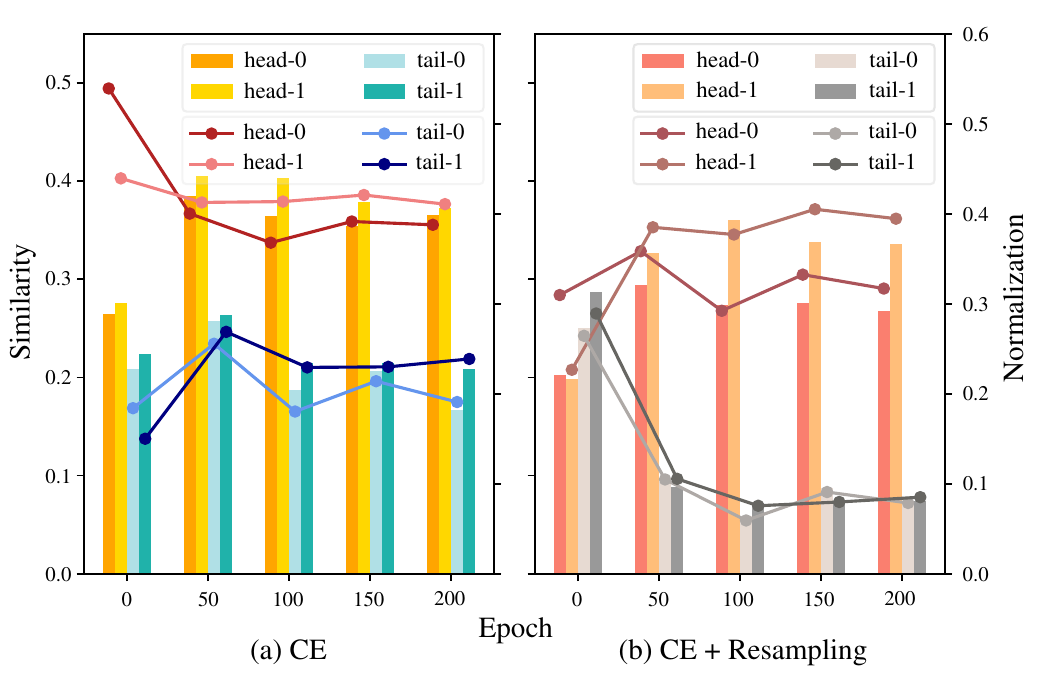}
	\vspace{-0.2in}
	\caption{Gradient imbalance in long-tail learning.
The bars denote the mean similarity between class-level and batch-level gradients in each batch. The dots represent the normalization of mean gradients of classes in epochs. We conduct the experiment on CIFAR10-LT, where (a) uses only cross entropy loss and (b) uses resampling strategy. We show the result of top two head classes and the last two tail classes.}
	\label{fig:sim}
	\vspace{-0.15in}
\end{figure}

In Fig.~\ref{fig:sim}, we measure the mean gradient similarity between class and whole gradient in each batch in different epochs. The similarity measures the contribution of gradients from different classes to the gradient descent process. A larger similarity means a larger contribution. 
Easy to observe in , \emph{the gradients of head and tail classes, which is presented as dots in the figure, are significantly imbalance}, where we have $\nabla_{\boldsymbol{\theta}}^\text{tail} < \nabla_{\boldsymbol{\theta}}^\text{head}$ in every epochs.
The reason of the gradient imbalance in long-tailed distribution, we argue, is the head-class samples make up the majority of most batches, resulting in the gradient domination of head classes in magnitude and direction over tail classes, which can be represented as $\nabla_{\boldsymbol{\theta}}^\top\nabla_{\boldsymbol{\theta}}^\text{head}>\nabla_{\boldsymbol{\theta}}^\top\nabla_{\boldsymbol{\theta}}^\text{tail}$. Finally, it cause the model could not obtain enough knowledge from tail-class data. At the same time, it raises confusion that the imbalance in Fig.~\ref{fig:sim}(a) is caused by the imbalanced data distribution in each bactch. We conduct a similar experiment with resampling strategy. 
As Fig.~\ref{fig:sim}(b) shows, the gradient imbalance phenomenon gets worse while using resampling strategy, which indicate the gradient imbalance is not caused by imbalanced distribution within each batch instead of the imbalanced distribution of the dataset.


To solve this problem, previous methods compensate tail-class gradients by raising the weight of tail-class loss by rebalancing strategies~\cite{lin2017focal,wu2020distribution}. 
However, as shown in Fig.~\ref{fig:seesaw}, intuitive rebalancing methods encounter the \emph{seesaw dilemma}, where the solutions may suffer from either overcompensation or undercompensation. 
Overcompensation refers to that the tail classes are overemphasized, while the head classes are underestimated, resulting in the learning of the head classes is excessively inhibited. 
Undercompensation is equivalent to no compensation.

In this paper, we seek to find an ideal compensation at each training iteration in long-tail learning, where the update should not damage any class in a long-tailed distribution.
To achieve this, for the first time to the best of our knowledge, we formulate the long-tailed learning into a multi-objective optimization problem as illustrated in the next subsection.


\begin{figure*}[t]
	\centering
	\includegraphics[width=1.0\linewidth]{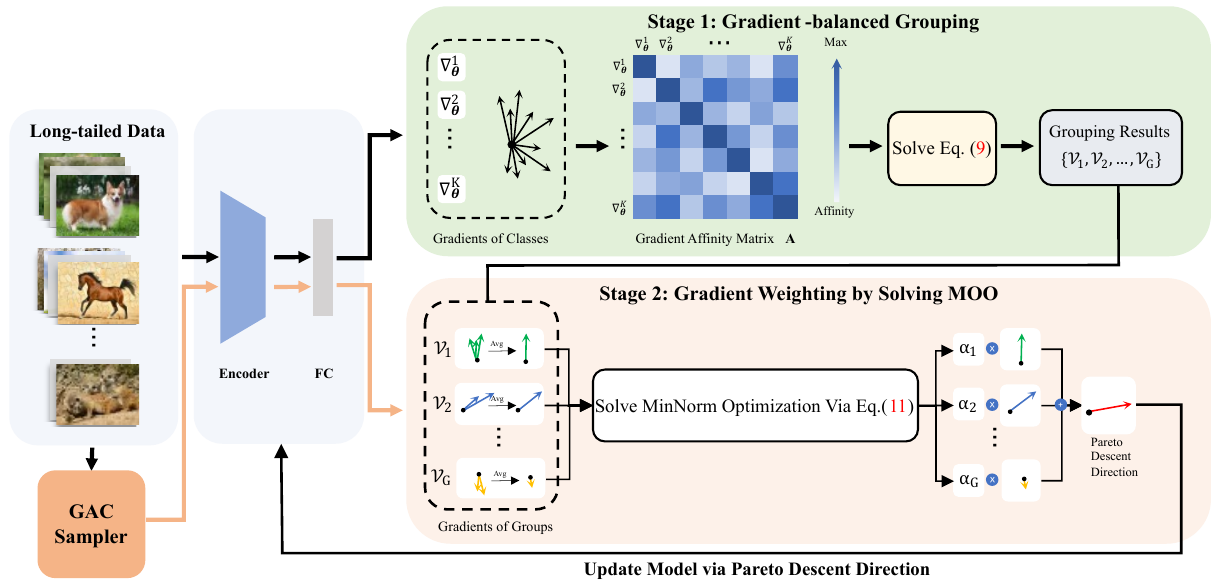}
	\vspace{-0.15in}
	\caption{Illustration of our proposed method.  At the first stage, we use GBG to gather the classes with high gradient similarity together. At the second stage, we use an averaging strategy to merge the gradients of the categories in the same group, and then solve a MOO problem to obtain an approximate Pareto descent direction in each iteration. 
    }
	\label{fig:methods}
	\vspace{-0.1in}
\end{figure*}

\subsection{LT Problem as Multi-Objective Optimization}



Multi-Objective Optimization (MOO) means optimizing multiple objectives simultaneously.
Given $T$ different objectives, a deep model with MOO yields the following multi-objective empirical risk minimization formulation: 

\begin{equation}
   \setlength{\abovedisplayskip}{5pt}
\setlength{\belowdisplayskip}{5pt}
\begin{aligned}
\mathop{\min}_{\boldsymbol{\theta} }\ \  \left\{L_i\left(\mc{D}_i\right),\cdots,L_T\left(\mc{D}_T\right)\right\},
\label{eq:pllbase}
\end{aligned}
\end{equation}
where $\mc{D}_i$ is the data of objective $i$.
Because of the conflict among objectives, the goal of MOO is to achieve Pareto optimality via training. 

\begin{definition}[\textbf{Pareto Optimality}]
        ~\\
      (1) (\emph{Pareto Dominate}) Let $\boldsymbol{\theta}_a$, $\boldsymbol{\theta}_b$ be two solutions for Problem~\eqref{eq:pllbase}, $\boldsymbol{\theta}_a$ is said to dominate $\boldsymbol{\theta}_b$ ($\boldsymbol{\theta}_a\prec\boldsymbol{\theta}_b$) if and only if $L_i(\boldsymbol{\theta}_a) \le L_i(\boldsymbol{\theta}_b)$, $\forall i \in \{1,2,\cdots,T\}$ and $L_i(\boldsymbol{\theta}_a) < L_i(\boldsymbol{\theta}_b)$, $\exists i \in \{1,2,\cdots,T\}$.\\
      (2) (\emph{Pareto Critical}) $\boldsymbol{\theta}$ is called Pareto critical if no other solution in its neighborhood can have better values in all objective functions.\\
      (3) (\emph{Pareto Descent Direction}) If $\boldsymbol{\theta}_1$ is not Pareto critical and can be updated to $\boldsymbol{\theta}_2$ by gradient $\mb{g}$. If $\boldsymbol{\theta}_2\prec\boldsymbol{\theta}_1$, say  $\mb{g}$ is a Pareto descent direction.
      \label{definition:1}
\end{definition}

 An MOO problem may have multiple solutions, consisting of a Pareto set, whose projection in loss space is called \emph{Pareto Front}. 
To approach the Pareto front in the loss space for all classes in a long-tailed distribution, we need to make each update under a Pareto descent direction that not damage any class's performance.
To this end, we convert the single objective loss function in Eq.~\eqref{eq:CE_batch} into a multi-objective optimization problem, which yields a gather of loss functions for each category
\begin{equation}
   \setlength{\abovedisplayskip}{5pt}
\setlength{\belowdisplayskip}{5pt}
\begin{aligned}
    \mc{L}\left(\boldsymbol{\theta};\mc{D}\right) 
    = \left\{{L}_1\left(\boldsymbol{\theta};\mc{D}_1\right),\cdots,{L}_{{K}}\left(\boldsymbol{\theta};\mc{D}_{{K}}\right)\right\},
    \label{eq:MOO}
    \end{aligned}
\end{equation}
where ${L}_k\left(\boldsymbol{\theta};\mc{D}_k\right)$ and $\mc{D}_k\subseteq \mc{D}$ represent loss function of class $k$ and the samples from class $k$ respectively. 

Let us review the seesaw dilemma under the MOO setting. 
After splitting training loss, we have $K$ different loss w.r.t. $K$ classes.
Subsequently, we obtain task-specific gradients $\{\nabla_1,\cdots,\nabla_K\}$ via derivation, where $\nabla_i=\nabla_{\boldsymbol{\theta}}^{L_i}$ and update only once.
That is, we need to aggregate all gradients into one.
A simple aggregation way is to set weights and sum class-level gradients.
At the iteration $n$, the problem can be reformulated as follows:
\begin{equation}
    \min_{\{\alpha_1,\cdots,\alpha_K\}}\quad   \left\{{L}_i\left(\boldsymbol{\theta}^{(n-1)}-\tau\sum\nolimits_{i=1}^K \alpha_i \nabla_i~;~\mc{D}_i\right)\Bigg|\forall i \right\}.
    \label{eq:moo2}
\end{equation}
To avoid the damage for all classes, we prefer to have ${L}_i\left(\boldsymbol{\theta}^{(n)};\mc{D}_i\right) \le {L}_i\left(\boldsymbol{\theta}^{(n-1)};\mc{D}_i\right)$ for any $i \in[1,K]$.
\emph{However, the multiple gradients may have large conflicts in terms of magnitude and direction.}
Inappropriate weighting may result in overcompensation and undercompensation that some classes may get decrease.
In contrast, the goal of multi-objective optimization is to achieve Pareto descent direction in each step, which will damage no class. 

Intuitively, using the loss of each category as optimization objectives can achieve better performance. However, in fact, more objectives do not necessarily mean better performance. Multi-objective optimization problems can pose significant challenges due to the increase in the dimensionality of search space  and the complexity of Pareto fronts as the number of objectives increases. 
Therefore, it is impractical to directly solve Problem~\eqref{eq:moo2} to achieve accurate Pareto descent direction, especially when the label space has large dimension.
Furthermore, the batch size is limited by the restricted size of hardware memory so it is difficult to cover all classes and store the gradients from all classes into the memory.
In next subsection, we propose a simple yet effective gradient-balancing grouping strategy to obtain an approximate Pareto descent direction.

\subsection{Gradient-based Class Grouping}



Class grouping~\cite{li2020overcoming} is one of effective solutions in long-tailed learning.
However, most of them rely on heuristic ideas and they can not guarantee a good compensation.
In this paper, we propose a Gradient-Balancing Grouping (GBG) strategy to solve the gradient conflict and obtain an approximate Pareto descent direction. 
GBG assigns classes with similar gradients direction into a group to make their gradients form a resultant force, which represents the approximate no-conflict direction to those of all corresponding classes in the group. 
Specifically, given class gradient in a batch $\{\nabla_1,\cdots,\nabla_c\}$, where $c$ denotes the contained class numbers of the batch. 
Let the category set be $\mc{C} = \{1,\cdots,K\}$.
We first compute a similarity matrix $\mathbf{A}$ to measure the similarity between any two gradients, and the element $\boldsymbol{A}_{i,j}$ is computed by
\begin{equation}
   \setlength{\abovedisplayskip}{5pt}
\setlength{\belowdisplayskip}{5pt}
\begin{aligned}
    \boldsymbol{A}_{i,j} = \frac{{\nabla_{\boldsymbol{\theta}}^i}{\nabla_{\boldsymbol{\theta}}^j}}{\Vert \nabla_{\boldsymbol{\theta}}^i\Vert \Vert\nabla_{\boldsymbol{\theta}}^j\Vert }.
    \label{eq:affinity_matrix}
    \end{aligned}
\end{equation}
According to the similarity matrix $\boldsymbol{A}$, we then build a graph $\mc{G} = \left\{\mc{V},\mc{E}\right\}$. 
$\mc{V}$ denotes the set of nodes in the graph, and each node represents a class.
$\mc{E} = \{A_{i,j}\}, i \leq j \leq K$ denotes the set of edges, where each edge represents the gradient similarity between class $i$ and class $j$. 
Our target is to find a way of grouping categories so that categories with high similarity on update habit are placed in a same group. Then, we define the affinity between groups as follows:
\begin{equation}
\setlength{\abovedisplayskip}{5pt}
\setlength{\belowdisplayskip}{5pt}
\begin{aligned}
    a\left(\mc{V}_m,\mc{V}_n\right) =\sum\nolimits_{i\in \mc{V}_m,j\in \mc{V}_n}\boldsymbol{A}_{i,j},
    \label{eq:group_affinity}
\end{aligned}
\end{equation}
where $\mc{V}_m, \mc{V}_n \subset \mc{V}$ represent two different groups and $\mc{V}_m\cap \mc{V}_n = \emptyset, \forall m \ne n $. 
Inspired by spectrum-clustering~\cite{ng2001spectral}, our target is equivalent to find a graph cutting $\mc{P}=\{\mc{V}_1,\mc{V}_2,\cdots,\mc{V}_G\}$ that minimizes the summation of affinity between groups, where $G$ is the number of groups. We formulate the problem as follows:
\begin{equation}
\setlength{\abovedisplayskip}{5pt}
\setlength{\belowdisplayskip}{5pt}
\begin{aligned}
\min\nolimits_{\mc{P}\in\mbb{P}}\quad\sum\nolimits_{\mc{V}'\in\mc{P}} a\left(\mc{V}', \mc{V}\right)-a\left(\mc{V}',\mc{V}'\right),
    \label{eq:graph_cut_optim}
\end{aligned} 
\end{equation}
where $\mbb{P}$ is the searching space for possible grouping strategy.
Then, we use NCut~\cite{shi2000normalized} to transform the optimization problem into the form of minimizing the Rayleigh entropy to obtain the partitioning result $\mc{P}$.


\par The grouping results $\mc{P}$ obtained through the above method ensure that the categories in the same group have similar update habit. Then, during the model training process, the gradients of each group we obtained in each batch is equivalent to the average of the gradients of the categories in their corresponding group, as shown in Fig.~\ref{fig:methods} (Stage 2). In other words, the gradients obtained from each group are as similar as possible to the gradients of any category within the group, which enables the gradients within each group to work together and implicitly increase the contribution of the tail class during training.

The overall class grouping procedure of GBG in LT problem is summarized in Algorithm~\ref{algorithm:1}. 
First, we fix the parameters of the initialized or pretrained model $f\left(\boldsymbol{\theta}\right)$. 
Next, we calculate the average gradients of each class and obtain the gradient similarity between each class through similarity function~\eqref{eq:affinity_matrix} to build an symmetrical gradient similarity matrix $\boldsymbol{A}$.
Eventually, we solve the graph cutting problem Eq.~\eqref{eq:graph_cut_optim} and get the final grouping result $\mc{P}$.
However, the gradient conflicts between  each group still exist.
In the following, we show how to solve the group-level MOO problem.
    
     
    
    

\begin{algorithm}[t]
 \renewcommand{\algorithmicrequire}{\textbf{Input:}}
 \newcommand{\LASTCON}{\item[\algorithmiclastcon]}
 \newcommand{\algorithmiclastcon}{\textbf{Output:}}
 \caption{Gradient-balanced grouping}
 \label{algorithm:1}
 \begin{algorithmic}[1]
  \REQUIRE Training set $\mc{D}$, Model parameters $\boldsymbol{\theta}$
  \FOR{ $k = 1 \rightarrow K$}
  \STATE Calculate average gradient ${\nabla_{\boldsymbol{\theta}}^k}$ of class $k$
  \ENDFOR \\
  \begin{small}
  // Compute gradient similarity matrix $\boldsymbol{A}$
  \end{small}
   
  \STATE $\bf{{\nabla}} \leftarrow[\nabla_{\boldsymbol{\theta}}^1,\cdots,\nabla_{\boldsymbol{\theta}}^K]^\top$
  \STATE $\boldsymbol{A} \leftarrow \left(\boldsymbol{{\nabla}}\cdot\boldsymbol{{\nabla}}^\top\right) / \{\Vert \boldsymbol{\nabla}\Vert\cdot \Vert\boldsymbol{{\nabla}}\Vert^\top\} $
  \STATE Build the graph $\mc{G} = \left(\mc{V},\mc{E}\right)$ according to $\boldsymbol{A}$ \\
    \begin{small}
  // Generate groups by solving Eq.~\eqref{eq:graph_cut_optim}
   \end{small}
  \STATE $\mc{P}^*\leftarrow \mathop{\arg\min}\limits_{\mc{P}\in\mbb{P}}\sum\limits_{\mc{V}'\in\mc{P}} a\left(\mc{V}', \mc{V}\right)-a\left(\mc{V}',\mc{V}'\right)$
  
  \LASTCON Group partitioning result $\mc{P}^*$
 \end{algorithmic}
\end{algorithm}



\subsection{Solving Group-level MOO Problem}

Following previous studies~\cite{zhou2020bbn}, our method is split into 2 stages as shown in Fig.~\ref{fig:methods}. At the first stage, we calculate the average gradients of each class and form a gradient similarity matrix. Then we divide categories into $G$ groups according to their gradient similarity. At the second stage, we bundle gradients of each group in each batch and form an MOO problem. Then, we solve the MOO problem to approximate a Pareto descent direction, achieving optimization of all groups at each iteration.


Based on the class grouping $\mc{P}^*=\{\mc{V}_1,\mc{V}_2,\cdots,\mc{V}_G\}$, the optimization goal of the long-tail problem can be converted into the training loss of $G$ groups
\begin{equation}
\setlength{\abovedisplayskip}{5pt}
\setlength{\belowdisplayskip}{5pt}
    \begin{aligned}
\min_{\boldsymbol{\theta}}\left({L}_1\left(\boldsymbol{\theta}\right),\cdots,{L}_G\left(\boldsymbol{\theta}\right)\right).
    \end{aligned}
\end{equation}
To efficiently solve the MOO problem, we adopts Multi Gradient Descent Algorithm (MGDA)~\cite{sener2018multi} that leverages the Karush-Kuhn-Tucker (KKT) conditions and transform the MOO problem into a min-norm single objective optimization as follows:
\begin{equation}
\setlength{\abovedisplayskip}{5pt}
\setlength{\belowdisplayskip}{5pt}
    \begin{aligned}
    \min_{\alpha_1,\cdots,\alpha_G} \quad &\left\| \sum\nolimits_{i=1}^G\alpha_i\nabla_{\boldsymbol{\theta}}{{L}_i\left(\boldsymbol{\theta}\right)} \right\|^2,\\
    \text{s.t.}\quad &\sum\nolimits_{i=1}^G\alpha_i=1 \text{~~and~~} \alpha_i\geq0,\forall i.
    \end{aligned}
    \label{eq:group_optim}
\end{equation}
As a min-norm single objective optimization, this problem can be easily solved by quadratic programming.
With the solution to this optimization problem, we obtain the final gradient for the long-tailed learning
\begin{equation}
  \setlength{\abovedisplayskip}{5pt}
\setlength{\belowdisplayskip}{5pt}
    \begin{aligned}
\boldsymbol{d}^*=\sum_{i=1}^G\alpha_i\nabla_{\boldsymbol{\theta}}{L}_i.
    \end{aligned}
\end{equation}
According to \citet{sener2018multi}, vector $\boldsymbol{d}^*$ is either zero or a feasible Pareto descent direction for all groups. 
We show the multi-objective optimization based gradient descent steps in Algorithm \ref{algorithm:2}.
Specifically, in every iteration, we compute the loss of each group and conduct backward over the model parameters for each loss to get $\nabla_{\boldsymbol{\theta}}{L}_i\left(\boldsymbol{\theta}\right)$. Then we acquire the weights $\{\alpha_1,\cdots,\alpha_G\}$ by solving~\ref{eq:group_optim} and use it carrying out weighted summation $\sum_{i=1}^G\alpha_i\nabla_{\boldsymbol{\theta}}{L}_i\left(\boldsymbol{\theta}\right)$ to get the final gradients  which are used to update model parameters.  
Moreover, we propose a simple but effective resampling method Group-Aware Completion (GAC) Sampler to guarantee that each batch contains samples from all groups, \ie, we have data from all groups in every mini-batch.
For each iteration, if the data of a group is missing in the mini-batch, we sample from the missing training data of the missing group with probability based on the class-balanced term~\cite{cui2019class}, so that the number of samples reaches 1/10 of the batch size. This can ensure that each batch contains samples from all groups and to some extent improve the contribution of tail-class samples.

\begin{algorithm}[t]
 \renewcommand{\algorithmicrequire}{\textbf{Input:}}
 \newcommand{\LASTCON}{\item[\algorithmiclastcon]}
 \newcommand{\algorithmiclastcon}{\textbf{Output:}}
 \caption{Update model by grouping MOO}
 \label{algorithm:2}
 \begin{algorithmic}[1]
  \REQUIRE Training set $\mc{D}$, Group results $\mc{P}^*$, Model parameters $\boldsymbol{\theta}$, Learning rate $\eta$
  \STATE Sample a mini-batch $\mc{B}=\{\mc{B}_1,\cdots,\mc{B}_G\}$ from the training set $\mc{D}$
  \FOR{ $i = 1 \rightarrow G$ }
  \STATE Compute group-level loss ${L}_i = {L}\left(\boldsymbol{\theta};\mc{B}_i\right)$
  \STATE Back-propaganda and compute gradients $\nabla_{\boldsymbol{\theta}}{L}_i$\
  \ENDFOR
  \STATE $\alpha_1^*,\cdots,\alpha_G^*\leftarrow$  Solve Eq.~\eqref{eq:group_optim}
  \STATE $\boldsymbol{\theta} \leftarrow \ \boldsymbol{\theta} - \eta\sum_{i=1}^G\alpha_i\nabla_{\boldsymbol{\theta}}{L}_i$
  \LASTCON Updated parameters $\boldsymbol{\theta}$
 \end{algorithmic}
\end{algorithm}
\par Compared with the original optimization problem containing a large number of objectives, our method greatly reduces the number of optimization objectives and is able to complete the training relatively efficiently. 

    
    

\section{Experiment}

\subsection{Datasets}
\noindent\textbf{CIFAR10/100-LT.}\ \ CIFAR10 with 10 classes and CIFAR100~\cite{krizhevsky2009learning} with 100 classes are balanced datasets, containing 50,000 training images and 10,000 validation images both. CIFAR10/100-LT are the long-tailed version of CIFAR10/100. Specifically, they are generated by downsampling CIFAR10/100 with different Imbalance Factor (IF) $\beta = {N}_\text{max}/{N}_\text{min}$ where ${N}_\text{max}$ and ${N}_\text{min}$ are the instance size of most frequent and least frequent classes in the training set~\cite{cui2019class,cao2019learning}. The validation set of CIFAR10-LT has 1,000 images per class and that of CIFAR100-LT is 100 images per class.

\noindent\textbf{ImageNet-LT.}\ \ Similar to long-tailed CIFAR, Liu et al.~\cite{liu2019large} proposed ImageNet-LT as the long-tailed version of original ImageNet. ImageNet-LT is sampled from vanilla ImageNet following a Pareto distribution with the power value $\alpha = 6$. It contains totally 115.8K training images of 1,000 categories with ${N}_{max} = 1,280$ and  ${N}_{min} = 5$. We use the balanced validation set of vanila ImageNet which contains 50 images per class.

\noindent\textbf{iNaturalist 2018.}\ \ iNaturailist 2018 is a large-scaled real-world dataset that naturally presents a long-tailed distribution. It consists of 437.5K images from 8,142 classes with $\beta = 512$. The validation set contains 24.4K images with 3 images per class to test our method.
\begin{table}[t]
  \centering
  \caption{Top-1 accuracy comparison on the CIFAR10/100-LT with IF = 100 and 50. $^{\dag}$ indicates the results reproduced by ourselves. The best and the second-best results are presented in \textbf{bold} and \underline{underline}, respectively. The indicates are the same in other tables. }
    \vspace{-0.1in}
    \resizebox{\linewidth}{!}{
    \begin{tabular}{l|cc|cc}
    \hline

    \hline

    {Method} & \multicolumn{2}{c|}{CIFAR10-LT} & \multicolumn{2}{c}{CIFAR100-LT} \\
    \hline
    {Imbalance Factor} & 100   & 50    & 100   & 50 \\
    \hline
    {CE} & 70.36 & 74.81 & 38.32 & 43.85 \\
    {CB-Focal~\cite{cui2019class}  } & 74.57 & 79.27 & 39.60  & 45.17 \\
    {LDAM-DRW~\cite{cao2019learning} } & 77.03 & -     & 42.04 & - \\
    {BBN~\cite{zhou2020bbn} } & 79.82 & 82.18 & 42.56 & 47.02 \\
    {SSD~\cite{li2021self} } & -     & -     & 46.00    & 50.50 \\
    {PaCo~\cite{cui2021parametric} } & -     & -     & 50.00   & 56.00 \\
    {RISDA~\cite{chen2022imagine} } & -     & -     & 50.16   & 53.84 \\
    {FDC~\cite{Ma2023FDC} } & 83.40     & 86.50     & 50.40   & 54.10 \\
    {DisVar~\cite{tian2023improving} } & 78.87     & 83.69     & 48.07   & 52.35 \\{BCL$^{\dag}$~\cite{zhu2022balanced} } & \underline{84.07} & \underline{86.98} & \underline{51.52} & \underline{56.23} \\
    \hline
    {GBG (Ours)} & \textbf{85.05} & \textbf{87.73} & \textbf{52.31} & \textbf{57.18} \\
    \hline

    \hline
    \end{tabular}%
    }
    \vspace{-0.1in}
  \label{tab:CIFAR}%
\end{table}%

\subsection{Experiment Setting}
We perform experiments on CIFAR10/100-LT with imbalance factor 100 and 50 and use ResNet-32 as backbone following ~\cite{cao2019learning,cui2019class}. For ImageNet-LT, we adopt ResNet-50 and ResNeXT-50 as our backbone. For iNaturalist 2018, we use ResNet-50 as our backbone. We set batch size as 256 for all datasets. We use fully-connected layer as classifier for all models. We train all above models on NVIDIA GeForce RTX 3090 GPU.
\subsection{Main Results}
We compare our method with state-of-the-art methods. We use top-1 accuracy as metric in all experiments. The comparison results are showed in Tables~\ref{tab:CIFAR}, \ref{tab:ImageNet} and \ref{tab:iNat}. 
Our method achieves the best performance compared with recent state-of-the-art methods on three benchmarks. 
On ImageNet-LT, our method gets 57.6\% in top-1 accuracy, which is 1.2\% over BCL with ResNet-50 and achieves 58.7\% which surpasses the second-best method for 1.8\%. The result on the iNaturalist 2018 shows that our method obtains 0.4\% over BCL. 
Because GBG gathers the classes with similar gradient direction, it makes the gradients of the groups be representative for those of classes within the group which implicitly increases the contribution of tail-class gradients. 


\begin{table}[t]
  \centering
  \caption{Top-1 accuracy comparison on ImageNet-LT.}
  \vspace{-0.1in}
  \resizebox{\linewidth}{!}{
    \begin{tabular}{l|c|c}
    \hline

    \hline

    {Method} & \multicolumn{2}{c}{ImageNet-LT} \\
    \hline
    {Backbone} & {ResNet-50} &{ResNext-50} \\
    \hline
    {CE} &  41.6 & 44.4 \\
    {Decoupling-LWS~\cite{kang2019decoupling}} &  - & 49.9 \\
    {RIDE(2 expert)~\cite{wang2020long}} &  54.4 & 55.9 \\
    {LADE~\cite{hong2021disentangling}} &  - & 51.9\\
    {Logit Adjustment~\cite{menon2020long}} &  51.1 & -\\
    {RISDA~\cite{chen2022imagine}} & 50.7 & - \\
    {Weight Decay~\cite{alshammari2022long}} & - & 53.9 \\
    {FDC~\cite{Ma2023FDC}} &   & 55.3\\
    {DisVar~\cite{tian2023improving}} &  49.4 & - \\
    {BCL$^{\dag}$~\cite{zhu2022balanced}} &  {\underline{56.4}} & {\underline{56.9}} \\
    \hline
    {GBG (Ours)} &  {\textbf{57.6}} & {\textbf{58.7}} \\
    
    \hline

    \hline
    \end{tabular}%
    }
    \vspace{-0.1in}
  \label{tab:ImageNet}%
\end{table}%
To gain deeper insights into the impact of our methods on various categories, we have partitioned the ImageNet-LT classes into three distinct subsets, following~\cite{zhu2022balanced}.
These subsets are characterized as Many ($>$100 images), Medium (20$\sim$100 images) and Few ($<$20 images) according to the instance number of classes 
The results presented in Table~\ref{tab:manyfew} unequivocally demonstrate notable performance enhancements across all three subsets. This observation underscores the effectiveness of our approach in elevating the performance of both tail and head categories. The MOO strategy employed in our method effectively addresses the challenge of imbalanced datasets, allowing for a harmonious training dynamic between the head and tail categories.

\begin{table}[t]
  \centering
  \caption{Top-1 accuracy comparison on iNatutalist 2018.}
   \vspace{-0.1in}
   \resizebox{0.9\linewidth}{!}{
    \begin{tabular}{l|c}
    \hline

    \hline

    {Methods} & {iNatualist 2018} \\
    \hline
    {CE} & {63.8} \\
    {LDAM-DRW~\cite{cao2019learning}} & {68.0} \\
    {BBN~\cite{zhou2020bbn}} & {69.6} \\
    {RIDE(2 experts)~\cite{wang2020long}} & {71.4} \\
    {LADE~\cite{hong2021disentangling}} & {70.0} \\
   {Logit Adjustment~\cite{menon2020long}} & {66.4} \\
    
    {RISDA~\cite{chen2022imagine}} & {69.2} \\
    {Weight Decay~\cite{alshammari2022long}} & {70.2} \\
    {DisVar~\cite{tian2023improving}} & {69.3} \\{BCL$^{\dag}$~\cite{zhu2022balanced}} & {\underline{71.5}} \\
    \hline
    {GBG (Ours)} & {\textbf{71.9}} \\
\hline

    \hline

    \end{tabular}%
    }
    \vspace{-0.1in}
  \label{tab:iNat}%
\end{table}%

\begin{table}[t]
  \centering
  \caption{Comparison on three subsets of ImageNet-LT.}
   \vspace{-0.1in}
   \resizebox{\linewidth}{!}{
    \begin{tabular}{l|cc|cc|cc|cc}
    \hline

    \hline

     {Methods (ResNext-50)} & \multicolumn{2}{c}{Many} & \multicolumn{2}{c}{Medium} & \multicolumn{2}{c|}{Few} & \multicolumn{2}{c}{All}\\
    \hline
     {$\tau$-norm~\cite{kang2019decoupling}} & \multicolumn{2}{c}{59.1} & \multicolumn{2}{c}{46.9} & \multicolumn{2}{c|}{30.7} & \multicolumn{2}{c}{49.4}\\
    {Balanced Softmax~\cite{ren2020balanced}} & \multicolumn{2}{c}{62.2}  & \multicolumn{2}{c}{48.8} & \multicolumn{2}{c|}{29.8} & \multicolumn{2}{c}{51.4} \\
    {Decoupling-LWS~\cite{kang2019decoupling}} & \multicolumn{2}{c}{60.2}  & \multicolumn{2}{c}{47.2}& \multicolumn{2}{c|}{30.3}& \multicolumn{2}{c}{49.9} \\
    {RIDE(4 Experts)~\cite{wang2020long}} & \multicolumn{2}{c}{\underline{68.2}} & \multicolumn{2}{c}{53.8} & \multicolumn{2}{c|}{36.0} & \multicolumn{2}{c}{56.8} \\
    {LADE~\cite{hong2021disentangling}} & \multicolumn{2}{c}{62.3} & \multicolumn{2}{c}{49.3} & \multicolumn{2}{c|}{31.2} & \multicolumn{2}{c}{51.9} \\
    {DisAlign~\cite{zhang2021distribution}} & \multicolumn{2}{c}{62.7} & \multicolumn{2}{c}{48.8} & \multicolumn{2}{c|}{31.6} & \multicolumn{2}{c}{51.8} \\

    {FDC~\cite{Ma2023FDC}} & \multicolumn{2}{c}{65.5} & \multicolumn{2}{c}{51.9} & \multicolumn{2}{c|}{\underline{37.8}} & \multicolumn{2}{c}{55.3} \\
     {BCL$^{\dag}$~\cite{zhu2022balanced}} & \multicolumn{2}{c}{66.9} & \multicolumn{2}{c}{\underline{54.3}} & \multicolumn{2}{c|}{{37.6 }} & \multicolumn{2}{c}{\underline{56.9}} \\
    \hline
    {GBG (Ours)} & \multicolumn{2}{c}{\textbf{69.6}} & \multicolumn{2}{c}{\textbf{55.8}} & \multicolumn{2}{c|}{\textbf{38.1}} & \multicolumn{2}{c}{\textbf{58.7}} \\
\hline

    \hline

    \end{tabular}%
    }
    \vspace{-0.1in}
  \label{tab:manyfew}%
\end{table}%


\subsection{Grouping Strategy Comparison}
In Table~\ref{tab:Grouping_strategy}, we compare our strategy with another two grouping rules, \ie, random grouping strategy and instance-numbers-based grouping strategy. Random grouping strategy partition the category set into 4 groups randomly.
We use several random seed to get different random grouping results and shows the average test results for fair comparison. For instance-numbers-based grouping strategy, we follow~\cite{li2020overcoming} to divide all categories into 4 groups according to the instance numbers of them, which means classes with similar instance number are in the same group.
The results show that our grouping strategy achieve the best performance among all grouping strategy, which prove the effectiveness of grouping via gradient.

\begin{figure}[t]
	\centering
\includegraphics[width=1.0\linewidth]{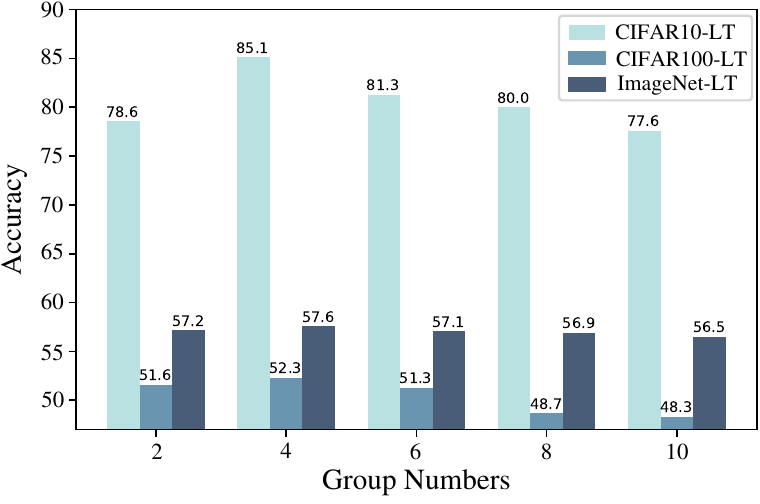}
	\vspace{-0.3in}
	\caption{Comparison results of different group numbers.} 
	\label{fig:group_numbers}
	\vspace{-0.1in}
\end{figure}

\begin{table}[t]
  \centering
  \caption{Grouping strategy comparison on ImageNet-LT (Avg$\pm$Std over 5 fixed seeds).}
   \vspace{-0.1in}
   \resizebox{0.7\linewidth}{!}{
    \begin{tabular}{l|c|c}
    \hline

    \hline
    
    {Strategy} & \multicolumn{2}{c}{ImageNet-LT} \\
    \hline
    {Backbone} & {ResNet-50} & {ResNext-50} \\
    \hline
   {Random} & {56.37$\pm0.41$}  & {56.96$\pm0.24$}\\
    {Instance number} & {56.61$\pm0.27$}  & {57.44$\pm0.38$}\\
    {GBG} & {{57.49$\pm0.29$}}  & {{58.61$\pm0.37$}}\\
    \hline

    \hline

    \end{tabular}%
    }
    \vspace{-0.15in}
  \label{tab:Grouping_strategy}%
\end{table}%

\subsection{Analysis on Different Group Numbers}
 In Fig.~\ref{fig:group_numbers}, we present the impact of different group numbers within our gradient-based class grouping mechanism. We conduct experiments across CIFAR10-LT, CIFAR100-LT, and ImageNet-LT datasets, varying the number of groups to determine the optimal configuration.
The outcomes of these experiments reveal that the use of four groups yields the most favorable results. This finding suggests that employing an excess number of objectives does not necessarily enhance performance in the MOO problems. Moreover, the strategy of assigning each class as an individual objective in CIFAR10-LT significantly undermines performance. This outcome is likely attributed to the heightened complexity of the optimization space when confronted with numerous objectives. The resulting increased likelihood of encountering local optima could lead to performance degradation.

\subsection{Ablation Study}
We perform several experiments to demonstrate the compacity and validity of our method. We split our methods into two parts including gradient-balanced grouping strategy (GBG) and multi-objective optimization (MOO) in Table.~\ref{tab:time}. We choose CIFAR10-LT (IF=100) for experiment and using ResNet-32 as backbone. We set the number of groups as four, and use average strategy to merge the gradients of groups. To validate the compatibility of proposed method, we add our method on naive model, Cross-Entropy (CE). The results show a significant improvement when our method is applied to CE, indicating its compatibility. 
\par By utilizing GBG, we implicitly enhance the impact of tail-class gradients. Therefore, GBG achieves a modest accuracy improvement of 0.13\% when adding GBG on BCL. On the other hand, directly employing the MOO method sets the loss of each category as an optimization objective. However, due to the limited batch size, data from tail classes are often absent in batches, resulting in the absence of certain optimization objectives in each iteration. Consequently, solely relying on MOO significantly decreases performance, which is only 74.58\%. In other words, the optimization objectives for the MOO problem are not fixed 
, considerably influencing the resolution of the MOO problem.

\par Our method combine Grouping and MOO. Through GBG, we put categories with high gradient similarity into a group, fixing the optimization objectives for MOO problem. We further solved the gradient conflicts between different groups through MOO, which enabled us to obtain approximate Pareto descent directions in every descent step. Building on this, we achieve a improvement of 0.85\% compared with only using Grouping.

\begin{table}[t]
  \centering
  \caption{Ablation study for our proposed method with BCL as baseline on CIFAR10-LT (IF = 100).}
   \vspace{-0.1in}
   \resizebox{0.5\linewidth}{!}{
    \begin{tabular}{l|c}
    \hline

    \hline
    
    {Method} & {Accuracy} \\
    \hline
    {CE} &  {71.36} \\
    {CE+GBG+MOO} &  {75.53} \\
    {BCL} &  {84.07} \\
    {BCL+GBG} &  {\underline{84.20}} \\
    {BCL+MOO} &   {74.58} \\
    {BCL+GBG+MOO} &  {\textbf{85.05}} \\
\hline

    \hline

    \end{tabular}%
    }
     \vspace{-0.2in}
  \label{tab:time}%
\end{table}%


\section{Conclusion}


In this paper, we found that gradient imbalance in the training process is a significant issue leading to poor performance in long-tailed learning. We thoroughly analyzed that the inappropriate compensation on the gradients of different classes resulting in the seesaw dilemma of previous methods. 
To solve the problem, we formulated the long-tailed recognition as an MOO problem and proposed a GBG algorithm to balance the gradient contributions of head and tail classes. 
Then, GBG makes classes with similar gradient directions form more representative gradients. 
With GBG, we approximately made every update of model parameters under a Pareto descent direction and provided ideal compensation to the tail classes. The experimental results on commonly used benchmarks proved that our method achieved new state-of-the-art performance, which demonstrated the superiority of our propose methods. 
In the future, we plan to further study how to adaptively adjust the number of groups and class grouping during training to reduce the time of hyperparameter tuning, enabling our method to be more efficiently applied to different datasets.

\bibliography{Reference}

\end{document}


\maketitle  


\appendix

\section{Group-Aware Completion Sampler}

In this section, we illustrate the detail of the GAC sampler.
The common random sampling cannot guarantee that each batch contains samples from all groups, in other words, it cannot ensure the invariance of the optimization objective in each batch. 
In GAC sampler, for each group $\mc{V}_i = \{c_1,\cdots,c_m\}$ where $c_m \in \mc{C}$, $m$ denotes the number of classes in group $\mc{V}_i$. We define the resampling probability $p_{j}$ for a sample from class $j$ in $\mc{V}_i$ as follows:
\begin{equation}
    \begin{aligned}
        p_{j} = \frac{1-{\beta}_j}{1-{\beta}_{j}^{N_{j}}},
    \end{aligned}
    \label{eq:resampling}
\end{equation}
where $N_{j}$ is the instance number of class $j$. To further increase the contribution of tail-class gradients, we make $\beta$ dependent on the degree of data imbalance in each group. Therefore, we calculate the $\beta_{j}$ as follows:
\begin{equation}
    \begin{aligned}
        \beta_j = \mu - \lambda \times \frac{N_j}{N_{min}},
    \end{aligned}
    \label{eq:e_normalize}
\end{equation}
where $N_{min}$ is the minimum instance number among the classes in group $\mc{V}_i$. $\mu$ and $\lambda$ are hyperparameters and we set them as $0.9999$ and $1e_{-7}$. Then, we  normalize the probability $p_{j}$ at group level as follows:
\begin{equation}
    \begin{aligned}
        p_{j} = \frac{p_j}{\sum\nolimits_{i=1}^mp_j
        }.
    \end{aligned}
    \label{eq:e_normalize}
\end{equation}

In specific, for each iteration, if data of group $\mc{V}_i$ is lacked in the mini-batch, we sample from the data of $\mc{V}_i$ with probability $p_{j}$, so that the number of samples reaches 1/10 of the batch size. This ensure the presence of data from all groups thus fixing the optimization objectives in each mini-batch.

\section{Gradient Distribution Analysis}
To demonstrate the effectiveness of our approach in improving the contribution of all classes, particularly the tail classes, we visualize the mean similarity between class-level and batch-level gradients in each mini-batch of CE and CE with GBG. 
\begin{figure}[H]
	\centering
	\includegraphics[width=1.0\linewidth]{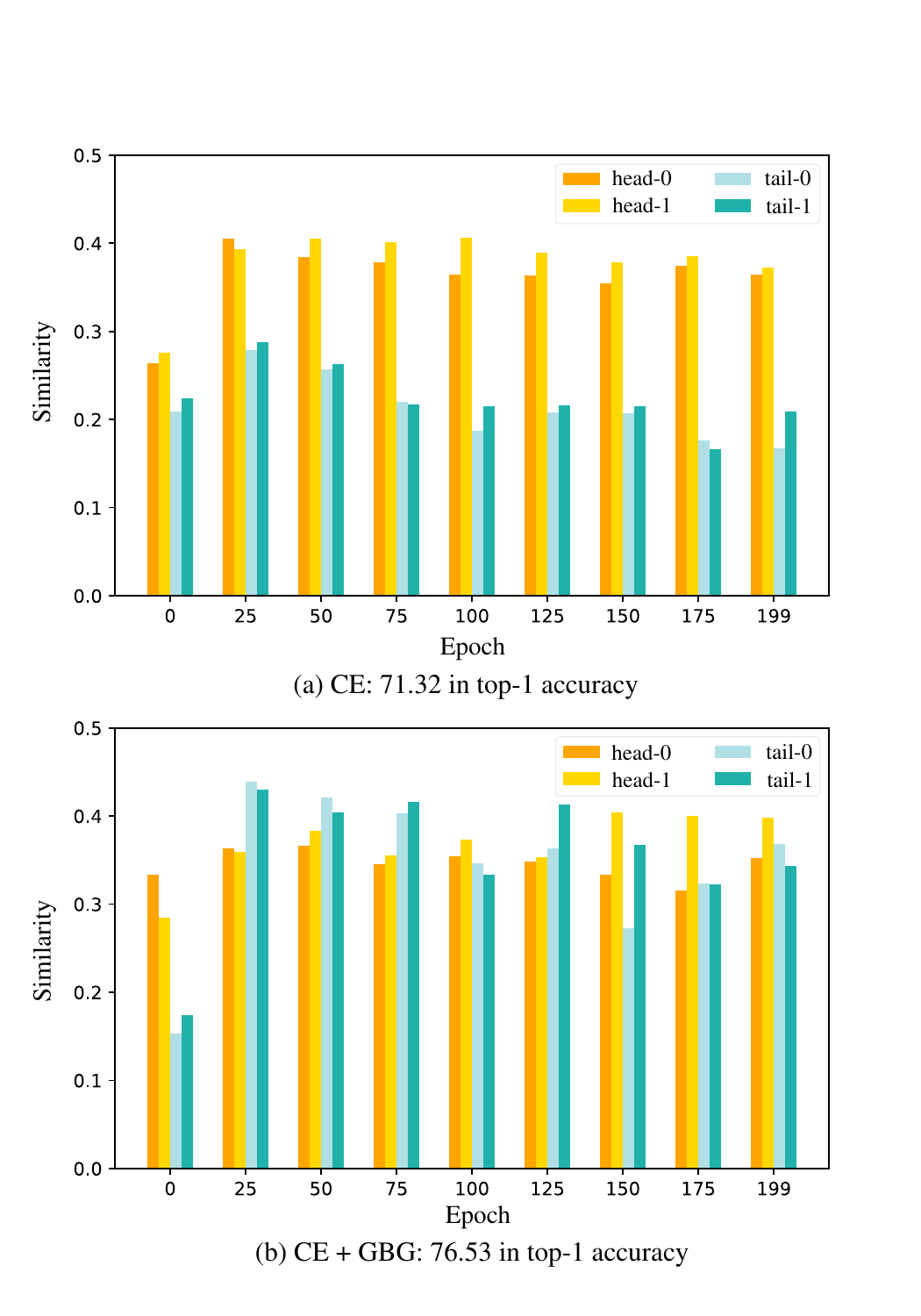}
	\vspace{-0.15in}
	\caption{Visualization of the mean similarity between class-level and batch-level gradients in every mini-batch. We show the result of naive model and our method respectively.}
	\label{fig:labelNfinal}
	\vspace{-0.15in}
\end{figure}
We perform this visualization experiment on CIFAR10-LT (IF=100), showcasing the two top head classes and two top tail classes. Our approach is added to the naive model, which utilizes ResNet-32 as the backbone, a linear classifier, and cross-entropy loss as the loss function. The comparison results are presented in Fig.~\ref{fig:labelNfinal}, where the bars represent the mean similarity that measures the contribution of gradients from different classes to the gradient descent process. A higher similarity indicates a greater contribution.
As shown in Fig.~\ref{fig:labelNfinal}, our approach improves the similarity of tail classes while negligibly reducing that of head classes. In other words, our approach can effectively increase the contribution of gradients for all classes during training, leading to improved performance across all classes.